\documentclass{edm_article}
\usepackage{subcaption}
\usepackage{booktabs}
\usepackage{xurl}
\usepackage[labelfont=bf]{caption}
\makeatletter
\def\addauthorsection{%
  \ifnum\originalaucount>6
    \edef\temp{\the\addauthors}%
    \ifx\temp\@empty
    \else
      \section{Additional Authors}\the\addauthors
    \fi
  \fi}
\makeatother
\begin{document}

\title{Utility-Preserving De-Identification for Math Tutoring: Investigating Numeric Ambiguity in the MathEd-PII Benchmark Dataset}

\numberofauthors{10}
\author{
Zhuqian Zhou\\
       \affaddr{Cornell University}\\
       \email{zz968@cornell.edu}
\and
Kirk Vanacore\\
       \affaddr{Cornell University}\\
       \email{kpv27@cornell.edu}
\and
Bakhtawar Ahtisham\\
       \affaddr{Cornell University}\\
       \email{ba453@cornell.edu}
\and
Jinsook Lee\\
       \affaddr{Cornell University}\\
       \email{jl3369@cornell.edu}
\and
Doug Pietrzak\\
       \affaddr{Fresh Cognate}\\
       \email{doug@freshcognate.com}
\and
Daryl Hedley\\
       \affaddr{Fresh Cognate}\\
       \email{daryl@freshcognate.com}
\and
Jorge Dias\\
       \affaddr{Fresh Cognate}\\
       \email{jorge@freshcognate.com}
\and
Chris Shaw\\
       \affaddr{UPchieve}\\
       \email{chris.shaw@upchieve.org}
\and
Ruth Schäfer\\
       \affaddr{Saga Education}\\
       \email{DrRuthSchafer@gmail.com}
\and
René F. Kizilcec\\
       \affaddr{Cornell University}\\
       \email{kizilcec@cornell.edu}
}
\sloppy

\maketitle

\begin{abstract}
Large-scale sharing of dialogue data is key to advancing the science of teaching and learning, yet rigorous de-identification remains a major barrier. In mathematics tutoring transcripts, numeric expressions frequently resemble structured identifiers (e.g., dates or IDs), leading generic Personally Identifiable Information (PII) detection systems to over-redact core instructional content and reduce data utility. This work asks how to detect PII while preserving educational utility, focusing on this “numeric ambiguity” problem. We introduce MathEd-PII, the first benchmark dataset for PII detection in math tutoring dialogues, built with human-in-the-loop LLM annotation. Using density-based segmentation, we show that false PII redactions cluster in math-dense regions, confirming numeric ambiguity as a key failure mode. We then compare four detection strategies: a Presidio baseline and three LLM-based approaches with basic, math-aware, and segment-aware prompting. Domain-aware prompting, including both math-aware (F1: 0.802) and segment-aware versions (F1: 0.821), substantially outperforms the baseline (F1: 0.379) while reducing numeric false positives, demonstrating that de-identification must incorporate domain context to preserve analytic utility. This work provides a new benchmark and evidence that utility-preserving de-identification for tutoring data requires domain-aware modeling.
\end{abstract}

\keywords{De-identification, PII Detection, Math Tutoring, Large Language Models, Privacy, Segmentation}

\bigskip
\bigskip
\section{Introduction}
The Educational Data Mining (EDM) community has long relied on large-scale open source data sets from digital learning platforms \cite{koedinger2010data,mihaescu2021review}. Many of these data sources can be easily de-identified because they are composed of action logs that, for the most part, do not typically contain personally identifiable information (PII). As more and more educational technologies provide dialogue-based tutoring at scale, either with humans or with artificial intelligence (AI) chatbots \cite{deacon2023impacts,henkel2024effective,prihar2022identifying,slijepcevic2025leveraging}, access to large corpora of tutoring conversations has become essential. However, sharing such data requires rigorous de-identification to remove direct identifiers (e.g., names and email addresses) and quasi-identifiers (e.g., gender, ages, locations, and school names) that could enable re-identification of specific individuals.

Recent work has begun to develop tools and workflows for de-identifying educational text \cite{Holmes2023StudentWriting, Holmes2023PIILO, Shen2025EnhancingDeid, Singhal2024EDMDeid, Zambrano2025LLMForumDeid, Zent2025PIIvot}. Yet specialized domains like mathematics tutoring introduce a distinct and underexplored challenge. In math dialogue, numbers are the primary medium of exchange. A teacher might ask, ``Are 4/12 and 2/6 the same?'' (fractions), or a student might calculate ``500 - 70 - 2000'' (subtraction). Standard PII detectors frequently misidentify these mathematical expressions as \texttt{DATE} or \texttt{US\_SSN} due to \textit{numeric ambiguity}, the overlap between the surface forms of mathematical content and sensitive identifiers. Over-identifying these math-related spans results in low precision, stripping the data of its core educational content and rendering it useless for downstream tasks.

A second barrier is the lack of domain-specific benchmarks. To date, no publicly available dataset exists for evaluating PII detection in math tutoring contexts, limiting the ability to rigorously study and improve de-identification methods for this increasingly important data modality.

To overcome these barriers, this work addresses the following research question (RQ): How can we accurately detect PII in math tutoring transcripts while preserving the educational utility of the data? We operationalize this question through three sub-questions that structure the remainder of the paper: \\
\textbf{RQ1 (Failure analysis):} To what extent does numeric ambiguity in mathematical discourse lead to false positive PII detection?\\
\textbf{RQ2 (Dataset construction):} How can a reliable benchmark for evaluating PII detection be created when original unredacted data cannot be shared?\\
\textbf{RQ3 (Method evaluation):} Do math-aware prompting strategies improve utility-preserving PII detection compared to domain-agnostic baselines?

To answer these questions, we begin with a large corpus of math tutoring transcripts that had already been redacted for PII by the data provider. The original unredacted data are unavailable because the data provider's internal data safety regulation requires responsible removal of PII before any data sharing can happen. Therefore, we developed a human-in-the-loop large language model (LLM) workflow to audit and repair PII redactions done by the data provider before they shared the data with us and to generate privacy-preserving surrogates. This process yields \textbf{MathEd-PII}\footnote{https://huggingface.co/datasets/NationalTutoringObservatory/MathEd-PII \label{fn:mathedpii}}, a benchmark dataset of 1,000 tutoring sessions (115,620 messages; 769,628 tokens) for evaluating PII detection in math tutoring dialogue.

We find that numeric expressions are a major source of false PII detections in math tutoring dialogue. Using a segmentation analysis to examine how detection errors differ between math-dense and conversational regions of transcripts, we observe 55.5\% of false redactions to occur in math-dense segments that cover only 36.6\% of the corpus, whereas true PII appears substantially less often in these regions (22.2\%). We evaluate four PII detection strategies: (i) a Presidio-based baseline, (ii) LLMs with basic prompting, (iii) LLMs with math-aware prompting, and (iv) LLMs with segment-aware prompting. Math-aware prompting substantially improves performance over the baseline (F1: 0.821 vs. 0.379), primarily by reducing numeric false positives. These findings demonstrate that de-identification for tutoring dialogue cannot be treated as a domain-agnostic preprocessing step; preserving analytic utility requires incorporating mathematical discourse context.

This paper makes three contributions aligned with the RQs above. We identify numeric ambiguity as a central failure mode in de-identifying math tutoring dialogue (RQ1). We then introduce MathEd-PII, the first benchmark dataset for PII detection in math tutoring transcripts (RQ2). Finally, we show that math-aware prompting substantially improves utility-preserving de-identification (RQ3).

\section{Related Work}

\subsection{Policy Context}
\label{sec:pc}
De-identification is a prerequisite for sharing educational interaction data at scale, but what counts as ``sufficient'' de-identification depends on the legal regime and the assumed adversary model. In the United States, the Children’s Online Privacy Protection Act (COPPA) and the Family Educational Rights and Privacy Act (FERPA) govern education records and only permits disclosure of de-identified information when there is no reasonable basis to believe the remaining information can be used to identify an individual \cite{FTC2013COPPA, PTAC2012DeidTerms}.

In parallel, the Health Insurance Portability and Accountability Act (HIPAA) provides two common pathways for de-identification of health information--\emph{Safe Harbor} (removal of 18 enumerated identifiers, e.g., names, geographic subdivisions smaller than a state, dates directly related to an individual, and unique identification numbers) and \emph{Expert Determination} (a documented, risk-based assessment) \cite{HHS2012DeidGuidance}. Although specific to health data, the 18 identifiers are widely used as a practical reference beyond that domain because they offer ``the promise of a straightforward application of rules, a repeatable process, and a known result: a dataset that is legally de-identified'' \cite[p.~24]{Garfinkel2015NISTIR8053}. Despite this apparent clarity, the National Institute of Standards and Technology (NIST) emphasizes that de-identification is inherently dynamic, framing it along an ``identifiability spectrum'' and cataloging transformation families such as suppression, generalization, perturbation, masking, and synthesis. Stronger privacy protections, it notes, typically come at the expense of analytic fidelity \cite{Garfinkel2015NISTIR8053,Garfinkel2023SP800188}.

\subsection{PII Detection in Unstructured Text}
\label{sec:piiunstructred}
Most practical pipelines for unstructured text de-identification begin with \emph{entity detection} (identifying spans that correspond to PII) and then apply an \emph{anonymization action} (e.g., redaction, masking, or surrogate replacement). Historically, detection approaches fall into three broad categories--rule-based, model-based, and LLM-based methods.

Rule-based PII detection systems rely on hand-engineered patterns and resources, including regular expressions for structured identifiers (e.g., emails, phone numbers, and dates), curated lexicons for names and locations, and deterministic context heuristics \cite{Neamatullah2008AutomatedDeid}. These methods can be highly precise for well-formed identifiers but brittle under conversational noise (misspellings, fragments, idiosyncratic formatting), and they require continual maintenance as domains shift \cite{Mishra2025FinancialHybrid}. 

A complementary line of work frames PII detection as a named entity recognition (NER) task, in which statistical or neural models predict entity spans over tokens or characters based on contextual cues. Contemporary NLP toolkits, most notably spaCy, provide pretrained and fine-tunable NER models that are commonly used as baseline PII detectors or adapted to domain-specific corpora \cite{Honnibal2020spacy}. In practice, many production systems combine \emph{rule-based} and \emph{model-based} paradigms to exploit their complementary strengths. Microsoft Presidio exemplifies this hybrid design by integrating rule-based recognizers for high-precision patterns with neural-network-based entity detectors within a unified anonymization framework \cite{MicrosoftPresidio}. 

More recently, LLMs have been applied to de-identification both via prompting and fine-tuning \cite{Shen2025EnhancingDeid, Singhal2024EDMDeid, Zambrano2025LLMForumDeid}. In prompting-based setups, the model is typically instructed (often with a schema) to tag or extract PII spans and entity types from raw text--either by returning structured outputs (e.g., JSON with character offsets and labels) or by emitting an annotated/redacted version of the transcript---whereas fine-tuning-based setups train the model on labeled examples to produce consistent token-level tags or span predictions under a fixed output format \cite{Shen2025EnhancingDeid}.

\subsection{Augmented and Synthetic data for PII Detection and Evaluation}
A practical challenge in PII detection research is the scarcity of richly annotated, shareable corpora, as releasing PII labels alongside text can increase re-identification risk. To address this, two complementary strategies are increasingly common: data augmentation and synthetic data generation.

Data augmentation for PII detection starts from de-identified or manually redacted text and inserts synthetic PII spans, often through surrogate replacement schemes such as ``hiding in plain sight,'' which preserve discourse coherence while reducing disclosure risk \cite{Carrell2013HIPS, Stubbs2015i2b2Overview}. In parallel, fully synthetic datasets have been constructed using template- or rule-driven generators and, more recently, LLMs conditioned on schema constraints (e.g., entity types, formats, and conversational style) \cite{MicrosoftPresidioResearch, Savkin2025SPY, Tang2023SyntheticClinicalTextMining}. These approaches offer controllability and improved coverage of rare entity types, but their effectiveness depends on the realism of the generated text: synthetic data that fails to capture domain-specific language, error patterns, or contextual ambiguity may yield overly optimistic performance estimates and limited transfer to real-world settings \cite{Savkin2025SPY, Tang2023SyntheticClinicalTextMining}.

\subsection{De-Identification of Educational Data}
Early work on de-identification in educational text focused on providing practical tooling and datasets to support privacy-preserving analysis of student writing and interaction data. Holmes et al.~\cite{Holmes2023StudentWriting, Holmes2023PIILO} introduced PIILO, an open-source system for labeling and obfuscating PII in educational corpora, and examined de-identification challenges in student-authored texts across digitally mediated learning environments. Building on this foundation, Singhal et al.~\cite{Singhal2024EDMDeid} conducted one of the first systematic evaluations of LLMs for educational data de-identification, using online course discussion forums as a testbed and demonstrating that LLMs substantially improve recall but often over-redact, motivating the need for utility-aware evaluation. Subsequent work by Zambrano et al.~\cite{Zambrano2025LLMForumDeid} expanded this analysis with larger-scale experiments and error analyses, further documenting precision--utility trade-offs in LLM-based anonymization of student forum posts. More recently, Shen et al.~\cite{Shen2025EnhancingDeid} proposed enhanced LLM-based pipelines that combine prompting and verification strategies to improve robustness across educational datasets, while Zent et al.~\cite{Zent2025PIIvot} introduced an anonymization framework tailored to question-anchored tutoring dialogues, emphasizing deployability and domain adaptation. Together, this line of work reflects a rapid shift from rule-based and toolkit-oriented approaches toward LLM-centric methods, alongside growing recognition that educational domains pose distinct challenges for balancing privacy protection and analytic utility as it is conversational and tightly interleaves task content with incidental personal references.

Math tutoring transcripts present an additional challenge: beyond conversational noise, they contain dense, non-sensitive numerals and operators whose surface forms overlap with common PII categories such as \texttt{AGE}, \texttt{DATE}, and \texttt{ID}. Over-redacting this math-related content would severely reduce the downstream educational and research-related utility of the dataset. However, numerical false positives in PII detection within mathematical contexts have received little attention. A major reason is the scarcity of labeled data: there are only two publicly available educational datasets with PII annotations, one consisting of student-authored articles submitted to a massive open online course (MOOC) on critical thinking through design\footnote{Available at https://www.kaggle.com/datasets/ langdonholmes/cleaned-repository-of-annotated-pii/data} \cite{Holmes2024CRAPII} and the other containing teacher–student chatroom messages for English language learning\footnote{Available via request at forms.gle/pKc48WMhnySC8zDk9} \cite{Caines2022TSCC}. To our knowledge, no public datasets exist for PII detection in math education contexts.

\section{Current Research Overview}

To answer the research questions outlined above, we follow a three-phase research workflow moving from dataset construction to analytical validation and finally to evaluation of math-aware de-identification strategies.

\textbf{Phase 1: Dataset Preparation} involves constructing MathEd-PII, a benchmark dataset for PII detection in math tutoring context. The construction includes identifying original over-redaction and establishing a clean ground truth through human-in-the-loop LLM-powered PII surrogation.

\textbf{Phase 2: Math Segmentation} focuses on developing a density-based module to isolate math-heavy regions, serving as an analytical lens to validate the hypothesis of numeric ambiguity that false PII redactions are structurally concentrated in mathematical content while true redactions are not.

\textbf{Phase 3: PII Detection and Evaluation} implements and compares four PII detection strategies—a Presidio-based baseline, LLMs with a basic prompt, a general math-domain-aware prompt, and a specific math-segment-aware prompt—evaluating Precision (P), Recall (R), and F1 overall, as well as stratified by PII types and by math versus non-math segments.

\section{Phase 1: Dataset Preparation}

Reference datasets for PII detection in math education are currently lacking. To enable rigorous evaluation, we constructed a benchmark dataset, \textbf{MathEd-PII}, from a PII-redacted large corpus.

\subsection{Source Corpus}
Our source corpus comprises 1,000 math tutoring sessions (115,620 messages; 769,628 tokens) from a U.S.-based tutoring platform. Among these, 4,648 (4\%) messages spread in 786 sessions feature 5,265 PII redactions applied by the platform's in-house PII detection system--Microsoft Presidio, which, as introduced in \ref{sec:piiunstructred}, is an industry-standard open-source software development kit (SDK) for PII detection by integrating rule-based recognizers for high-precision patterns with neural-network-based entity detectors within a unified anonymization framework \cite{MicrosoftPresidio} with customizable rules. The provider redacts 12 types of PII, such as \texttt{PERSON}, \texttt{LOCATION}, \texttt{US\_DRIVER\_LICENSE}, etc. The full list of PII types can be found in Table~\ref{tab:segment_results}.

Initial inspection suggests that several redaction types are not consistently reliable indicators of true PII in math tutoring dialogue. In particular, some types that resemble structured numeric identifiers (e.g., \texttt{COURSE\_NUMBER} and \texttt{US\_DRIVER\_LICENSE}) are frequently triggered by math content rather than identity disclosure (see an example of incorrect redactions and inferred original text\footnote{Original un-redacted text is not available due to the data provider's privacy compliance. The original text here is inferred by an LLM based on the context. Please see Section~\ref{sec:method} for more details.} in Table~\ref{tab:quote}). Other types (notably \texttt{LOCATION}, \texttt{NRP}\footnote{A person’s Nationality, Religious or Political group.\label{fn:nrp}}, and \texttt{URL}) appear mixed, containing both correct redaction and substantial over-redaction.

\begin{table}[h]
    \caption{\textbf{Incorrect PII Redactions in the Source Corpus and Inferred Original Text}}
    \label{tab:quote}
    \centering
    \begin{tabular}{p{0.45\linewidth} p{0.45\linewidth}}
    \toprule
    \textbf{Redacted Message} & \textbf{Inferred Original Message} \\ \midrule
    9351 / 7 = 1335 \newline <US\_DRIVER\_LICENSE> & 9351 / 7 = 1335.8571 \\
    go from radians to degrees \newline it would \newline <COURSE\_NUMBER>/pi & go from radians to degrees it would 180/pi \\
    so it would be 1.15x=<PHONE\_NUMBER> & so it would be 1.15x= 368000 \\
    \bottomrule
    \end{tabular}
\end{table}

\subsection{Method: Human-In-The-Loop PII Redaction Evaluation and Surrogate Generation}
\label{sec:method}

\begin{figure*}[t]
    \centering
    \includegraphics[width=1\linewidth]{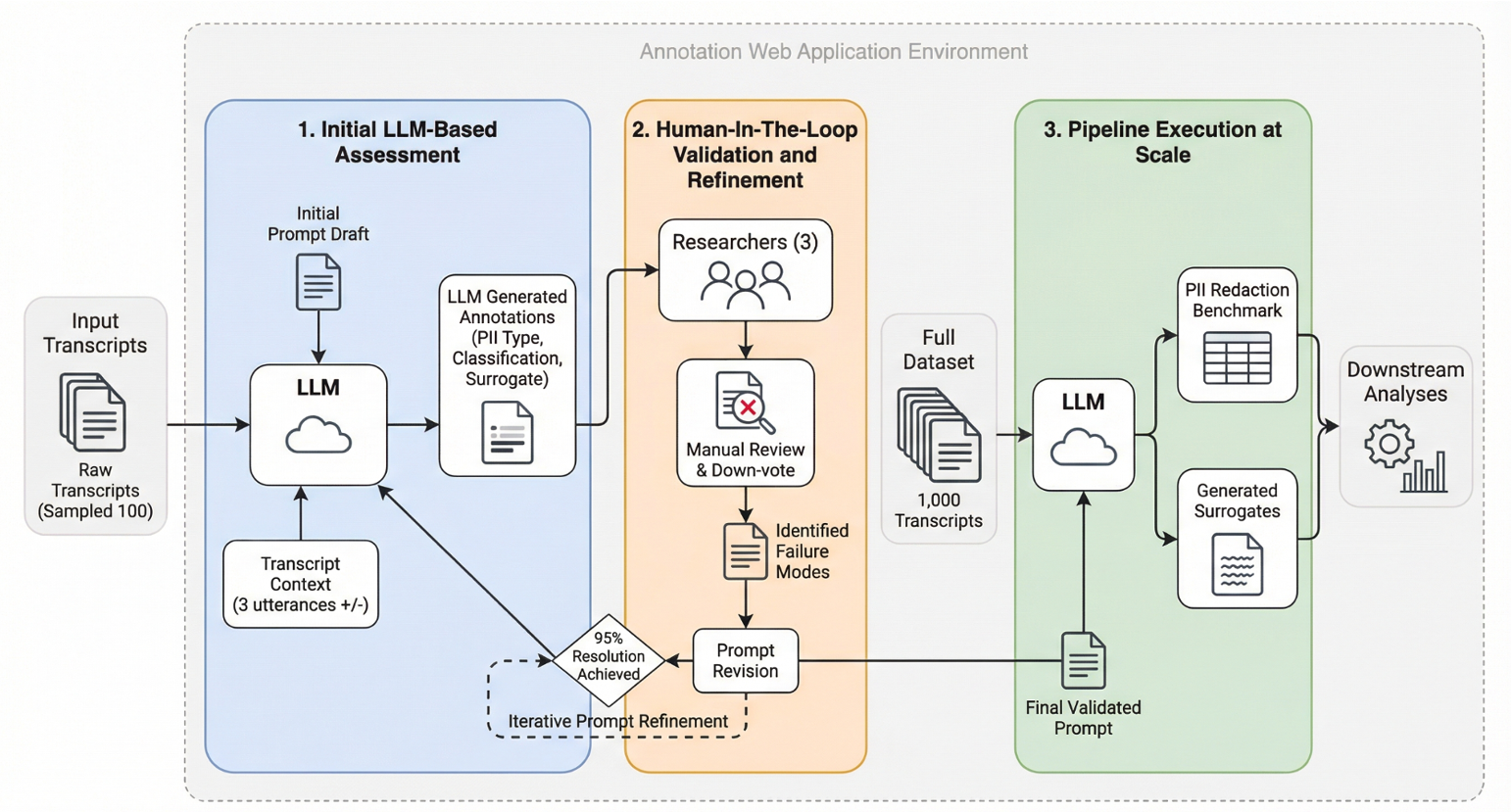}
    \caption{\textbf{The Three-Step Human-In-The-Loop MathEd-PII Benchmark Dataset Preparation Workflow.}}
    \label{fig:workflow}
    \Description{Workflow diagram for an annotation web application environment showing a three-stage pipeline for LLM-assisted transcript annotation and validation. On the far left, a stack of “Input Transcripts” labeled “Raw Transcripts (Sampled 100)” feeds into Stage 1, titled “Initial LLM-Based Assessment,” shown in a blue panel. Within this stage, an “Initial Prompt Draft” and “Transcript Context (3 utterances +/-)” are provided to an LLM, which generates annotations for “PII Type, Classification, Surrogate.” The generated annotations are passed to Stage 2, “Human-In-The-Loop Validation and Refinement,” displayed in an orange panel. Three researchers manually review and down-vote incorrect outputs, identify failure modes, and revise prompts. A decision diamond labeled “95\% Resolution Achieved” indicates iterative prompt refinement between the LLM assessment and prompt revision steps. The finalized validated prompt is then used in Stage 3, “Pipeline Execution at Scale,” displayed in a green panel. Here, the LLM processes a “Full Dataset” of 1,000 transcripts to produce outputs including a “PII Redaction Benchmark” and “Generated Surrogates,” which feed into “Downstream Analyses.” The entire workflow is enclosed within a dashed border labeled “Annotation Web Application Environment.”}
\end{figure*}

To evaluate the extent of incorrect redactions and construct a reliable benchmark for PII detection from the current dataset, we employed a human-in-the-loop LLM workflow to assess the quality of existing PII labels and generate plausible surrogates, thereby restoring transcript completeness for downstream investigation. This design choice is motivated by three practical constraints. First, the original un-redacted transcripts are unavailable due to privacy and data-protection requirements, precluding direct comparison against ground-truth PII. Second, manually reviewing and reconstructing every redaction at scale would be prohibitively costly and time-consuming. Third, LLMs are well-suited for this task because they can leverage rich contextual information to infer the most plausible lexical realizations consistent with surrounding discourse, enabling systematic approximation of redacted content without reintroducing real personal identifiers. Human oversight is incorporated throughout the workflow to validate model outputs, refine prompts, and ensure that generated surrogates remain contextually coherent while preserving privacy. This workflow consists of the following three steps and is illustrated in Figure~\ref{fig:workflow}.

First, \textbf{Initial LLM-Based Assessment}: We first drafted an initial prompt that instructed the LLM to evaluate messages from a randomly sampled set of 100 transcripts, using a context window of at least three utterances preceding and following each target message. For each message, the LLM was prompted to identify any redacted PII or PII missed by the data provider's internal redaction process, determine its PII type, and classify the redaction as ``PII'', ``Not PII'', or ``Uncertain'' based on contextual evidence. When PII was identified, the LLM was further instructed to generate context-appropriate surrogate text that preserved semantic coherence while remaining substantially different from any original PII, when available.

Second, \textbf{Human-In-The-Loop Validation and Refinement}:  Three researchers—the first three authors of this paper—then manually reviewed the LLM outputs. Each researcher was assigned one-third of the outputs and down-voted instances in which any component of the output was deemed incorrect, including the identified PII type, quality classification, or the generated surrogate. Based on these annotations, the prompt was revised iteratively to address observed failure modes. This process continued until at least 95\% of previously down-voted issues were successfully resolved by the updated prompt.

Third, \textbf{Pipeline Execution at Scale}: The final, validated prompt\footnote{see 1 in https://github.com/ZhuqianZhou/MathEd-PII/blob/69012c0dbde3b3976f36533e2e44f5a5bfdfd718/ EDM2026\_paper\_183\_Appendix.pdf} for redaction evaluation and surrogate generation was subsequently applied to the full 1,000-transcript dataset to produce the benchmark used in downstream analyses.

All three steps were conducted using an open-sourced annotation web application, Sandpiper\footnote{https://sandpiperresearch.org/}, optimized for stable, human-in-the-loop annotation iteration. The system provides a robust, interactive user interface that allows researchers to directly review, up-vote, or down-vote AI-generated annotations, enabling efficient human verification and correction at scale. Its modular architecture and worker-based execution model ensure stable handling of large annotation workloads and repeated LLM inference, supporting consistent evaluation across iterations.

\subsection{Source Corpus Evaluation Results}

Applying the procedures described above, we discovered both recall and precision issues of the data provider's internal PII redaction presented in the source corpus as described below.

\textbf{The Recall Issue}: The human-in-the-loop LLM workflow discovered 201 messages containing PII that the data provider's system failed to redact, spanning \texttt{PERSON} (159), \texttt{GRADE\_LEVEL} (23), \texttt{AGE} (8), \texttt{DATE} (4), \texttt{COURSE\_NUMBER} (3), \texttt{NRP}\footnote{see foot note~\ref{fn:nrp}} (3), \texttt{SCHOOL} (2), and \texttt{URL} (1). Notably, \texttt{DATE} and \texttt{AGE} were newly introduced PII types under the Safe Harbor framework (see Section~\ref{sec:pc}) and were not included among the original dataset’s 12 PII categories. Still, this demonstrates a substantial recall issue: a widely used approach to de-identification, Microsoft Presidio, failed to identify 159 names in tutoring transcripts. Table~\ref{tab:missedPerson} presents simulated examples of missed names. These examples were generated from failure patterns observed in the original dataset and do not refer to any real people. Readers can replicate the same failure using Presidio's default Analyzer Engine with customized rules.

\begin{table}[h]
    \caption{\textbf{Simulated Examples of \texttt{PERSON} that Presidio Failed to Detect and Similar Cases that Presidio Successfully Detected}}
    \label{tab:missedPerson}
    \centering
    \begin{tabular}{ll}
    \toprule
    \textbf{Unsuccessful Detection} & \textbf{Successful Detection} \\ \midrule
    hey olivia & hey Olivia \\
    Hi Alex & Hi Alex, \\
    I am Lam & i am Lam \\
    \bottomrule
    \end{tabular}
\end{table}

\textbf{The Precison Issue}: Of the 4,648 originally redacted messages, 3,015 (64.87\%) contained redactions evaluated as ``Not PII'', indicating low precision in their internal system. Several PII types contained predominantly false positives, with the following four types receiving more than 90\% ``Not PII'' flags--\texttt{US\_DRIVER\_LICENSE} (99.36\%), \texttt{COURSE\_NUMBER} (96.65\%), \texttt{PHONE\_NUMBER} (93.33\%), and \texttt{NRP} (90.64\%), the first three of which are all numeric identifiers further corroborating our initial findings. Table~\ref{tab:quote} presents some examples of such over-redaction.

\subsection{The MathEd-PII Surrogate Benchmark}

By replacing all of the originally redacted PII and newly identified PII with LLM-generated, context-aware surrogates and removing PII annotations for those that were labeled as ``Not PII'', we obtained the MathEd-PII surrogate benchmark dataset. A comparison of basic statistics for the source corpus and MathEd-PII is presented in Table \ref{tab:stats_comparison}.

\begin{table}[h]
\caption{\textbf{PII Statistics Comparison between Source Corpus and MathEd-PII (ordered by the most common PII to the least in MathEd-PII)}}
\label{tab:stats_comparison}
\centering
\begin{tabular}{lrr}
\toprule
\textbf{Category} & \textbf{Source Corpus} & \textbf{MathEd-PII} \\
\midrule
Transcripts & 1,000 & 1,000 \\
Messages & 115,620 & 115,620 \\
PII Labels (Total) & 5,263 & 1,995 \\
\midrule
PERSON & 1,915 & 1,424 \\
URL & 245 & 187 \\
LOCATION & 595 & 121 \\
GRADE\_LEVEL & 87 & 107 \\
SCHOOL & 88 & 73 \\
COURSE\_NUMBER & 1,103 & 40 \\
NRP & 235 & 25 \\
AGE & 0 & 8 \\
DATE & 0 & 4 \\
US\_DRIVER\_LICENSE & 941 & 2 \\
PHONE\_NUMBER & 30 & 2 \\
IP\_ADDRESS & 2 & 2 \\
US\_BANK\_NUMBER & 20 & 0 \\
US\_PASSPORT & 2 & 0 \\
\bottomrule
\end{tabular}
\end{table}

\section{Phase 2: Math Segmentation and Numeric Ambiguity}

Since major ``Not PII'' redactions are numeric identifiers, a natural explanation is \emph{numeric ambiguity}: mathematical content in tutoring transcripts often shares surface-level similarities with numeric PII, making the two difficult to distinguish. Assessing the extent to which numeric ambiguity degrades PII detection—and developing effective mitigation strategies—therefore requires distinguishing math-dense regions from non-mathematical conversational text.

\subsection{Method: Math Density, Semantic Similarity, and Segmentation Optimization}
To distinguish math-focused sections in a transcript, we applied a segmentation strategy by first identifying anchor math messages with high density of math vocabulary and then expanding the segment based on semantic similarity \cite{galley2003discourse, Huang2015BiLSTMCRF, misra2011text}.

The construction of the math vocabulary set was informed by a comprehensive list of 2,824 K-12 math learning components\footnote{https://docs.learningcommons.org/knowledge-graph/understanding-knowledge-graph/about-knowledge-graph} curated by the Learning Commons initiative and the Achievement Network \cite{learningcommons_learning_components}, which are granular representations of individual math skills and concepts encompassing the Common Core State Standards for Mathematics (CCSSM) and 18 U.S. state math standards (e.g., California, New York, Texas, and Washington). Some examples of the learning components are ``Use conversions to solve multi-step real-world problems'', ``Determine the lateral surface area of three-dimensional cylinders in real-world problems'', and ``Represent proportional relationships with equations''. The math vocabulary used for math density calculation includes a list of math terms derived from these learning components and regular expressions (RegEx) of common math expressions. The vocabulary can be found here\footnote{see 2 in https://github.com/ZhuqianZhou/MathEd-PII/blob/69012c0dbde3b3976f36533e2e44f5a5bfdfd718/ EDM2026\_paper\_183\_Appendix.pdf}.

The \textbf{Math Density ($D_{math}$)} of a message is a weighted sum of single-word math vocabulary terms ($V_{math}$), multi-word phrases ($P_{phrase}$), and RegEx patterns ($R_{pattern}$):

\begin{equation}
\resizebox{0.9\hsize}{!}{$D_{math}(m) = \frac{1}{|W|} \left[ \sum\limits_{w \in W} \mathbb{I}(w \in V) + 1.5 \sum\limits_{p \in P} C(p, m) + 2.0 \sum\limits_{r \in R} C(r, m) \right]$}
\end{equation}

where $W$ is the set of tokens in the message, $\mathbb{I}(\cdot)$ is the indicator function, $C(\cdot, m)$ counts Occurrences in message $m$. The weights were empirically set to attribute higher significance to multi-term concepts and structural math patterns. 

A message exceeding a threshold of Math Density, \textit{Anchor Threshold} ($T_{anchor}$) at 0.05, becomes an anchor message and initializes a math segment. Adjacent messages are added to the math segment if their cosine similarity to the segment centroid (the aggregated semantic meaning of the math context accumulated in the segment computed via \texttt{all-MiniLM-L6-v2} embeddings) exceeds a \textit{Similarity Threshold} ($T_{sim}$) at 0.3. The thresholds were determined via a grid search optimization process around the average math density of the dataset (0.07) from 0.05 to 0.10 in increments of 0.01 and a range of similarity threshold from 0 to 0.5 with a step size of 0.1 with the objective to maximize the capture of ``Not PII'' (false positives) and minimize the capture of ``PII'' (true positives) by the math segments. Notably, across all threshold configurations evaluated in the grid search, the proportion of false positives consistently exceeded the proportion of true positives captured by the math segments. This indicates that the finding elaborated below is robust to the choice of threshold. The optimization process and analysis can be seen here\footnote{see 3 in https://github.com/ZhuqianZhou/MathEd-PII/blob/69012c0dbde3b3976f36533e2e44f5a5bfdfd718/ EDM2026\_paper\_183\_Appendix.pdf}. 

\subsection{False Positive PII in Math Segments and Numeric Ambiguity}

\begin{figure*}[t]
    \centering
    \includegraphics[width=1\linewidth]{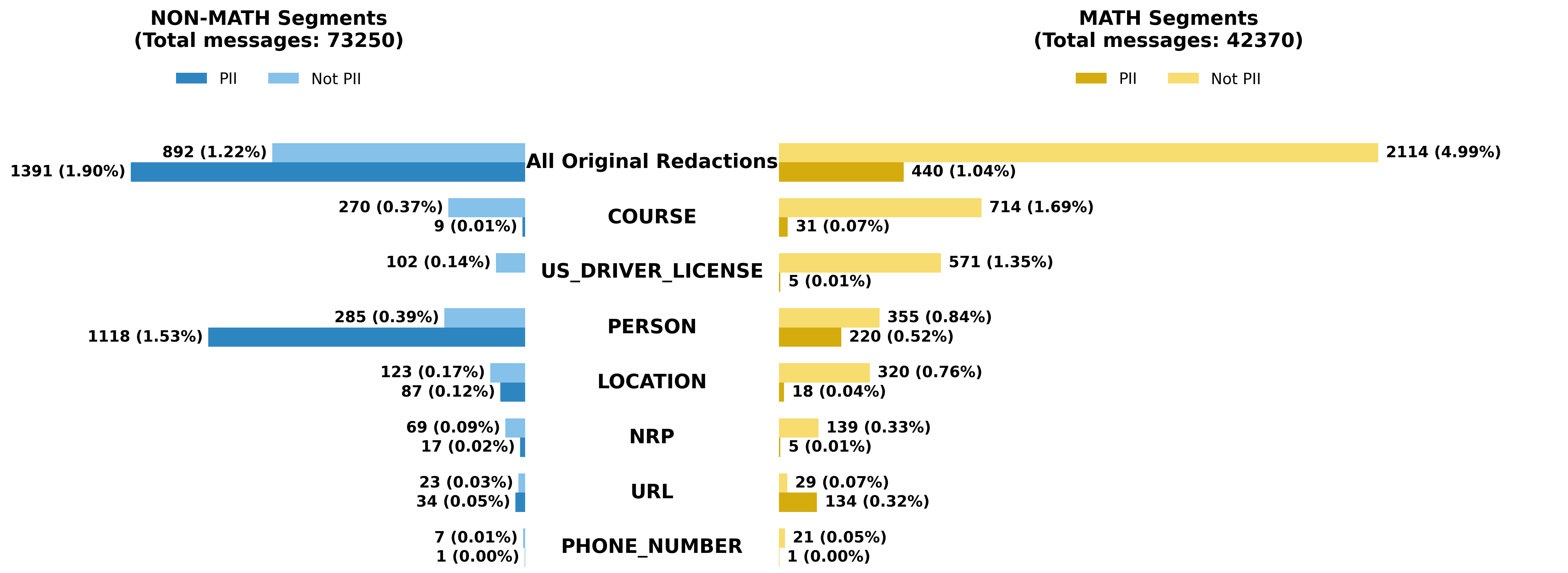}
    \caption{\textbf{Distribution of Original PII Redactions from the Data Provider's In-House System (Presidio and Customized Rules) across Different PII Types and Math and Non-Math Segments in terms of true PII and false positives (``Not PII''). Types are ordered by the frequency of ``Not PII'' in Math Segments. The first seven types are presented.}}
    \Description{Mirrored horizontal bar chart comparing the distribution of personally identifiable information (PII) and non-PII entities in NON-MATH versus MATH transcript segments. The left side represents “NON-MATH Segments (Total messages: 73,250)” using blue bars, while the right side represents “MATH Segments (Total messages: 42,370)” using yellow bars. Each category displays counts and percentages for PII and Not PII occurrences. Categories listed vertically in the center are: “All Original Redactions,” “COURSE,” “US_DRIVER_LICENSE,” “PERSON,” “LOCATION,” “NRP,” “URL,” and “PHONE_NUMBER.”  For NON-MATH segments, “All Original Redactions” contains 1,391 PII instances (1.90\%) and 892 non-PII instances (1.22\%). “PERSON” has 1,118 PII instances (1.53\%) and 285 non-PII instances (0.39\%). “LOCATION” includes 87 PII and 123 non-PII mentions. “URL” is the only category where PII exceeds non-PII aside from PERSON, with 34 PII versus 23 non-PII instances. For MATH segments, “All Original Redactions” contains 440 PII instances (1.04\%) and 2,114 non-PII instances (4.99\%), indicating substantially more false-positive-style non-PII content. “COURSE” and “US_DRIVER_LICENSE” categories show especially high non-PII counts relative to PII, with 714 versus 31 and 571 versus 5 respectively. “PERSON” has 220 PII and 355 non-PII instances. The chart emphasizes that math-related segments contain many more ambiguous non-PII terms that resemble PII categories compared with non-math segments.}
    \label{fig:pii_dist}
\end{figure*}

Figure \ref{fig:pii_dist} illustrates the distribution of PII categories across mathematical and non-mathematical segments, distinguishing between true positives (validated PII) and false positives (Not PII) arising from numeric ambiguity. 

It reveals a stark contrast in detection accuracy between the two domains. In math segments, the data provider's in-house detection system produced 2,539 false positives compared to only 520 true PII instances---a false-to-true ratio of approximately 4.88:1. In contrast, non-math segments exhibited a much higher precision, with 929 false positives and 1,455 true PII instances (a ratio of 0.64:1). 

Specific numeric categories were nearly entirely composed of false positives in mathematical contexts. For instance, as shown in the MATH Segment of Figure \ref{fig:pii_dist}, categories such as \texttt{DATE}, \texttt{US\_DRIVER\_LICENSE}, and \texttt{LOCATION} contain substantial numbers of false positives. Interestingly, the \texttt{PERSON} category also presents a high number of false positives within math segments. Qualitative review indicates that these names are frequently tied to mathematical word problems (e.g., ``If Alex has 5 apples...'') rather than the actual identities of the student or tutor. While over-redacting these fictional entities is perhaps less detrimental to educational utility than over-redacting the mathematical operations themselves, it further illustrates the inability of domain-agnostic systems to distinguish between the \textit{context} of a name (math problem vs. personal disclosure). 

This empirical evidence confirms that numeric ambiguity is a structural failure mode in de-identifying math tutoring transcripts: the surface forms of mathematical expressions and problem-related entities systematically overlap with structured personal identifiers, leading generic detection systems to over-redact critical instructional content.

\section{Phase 3: PII Detection and Evaluation}

\subsection{PII Detection Methods}

\subsubsection{Baseline: Microsoft Presidio}
As a baseline, we deployed Microsoft Presidio (v2.2) \cite{MicrosoftPresidio}. We utilized its default analyzer which orchestrates a set of predefined recognizers. For high-structure entities (e.g., \texttt{EMAIL\_ADDRESS}, \texttt{URL}, \texttt{IP\_ADDRESS}, \texttt{PHONE\_NUMBER}, \texttt{US\_SSN}, \texttt{US\_DRIVER\_LICENSE}), Presidio employs regular expressions (RegEx). For context-dependent entities (e.g., \texttt{PERSON}, \texttt{LOCATION}, \texttt{NRP}), it leverages a spaCy-based NER model (\texttt{en\_core\_web\_lg} and \texttt{en\_core\_web\_trf} used here). To address domain-specific requirements, we implemented custom RegEx-based recognizers for \texttt{SCHOOL}, \texttt{COURSE\_NUMBER}, and \texttt{GRADE\_LEVEL}. We mapped Presidio's output entities to the 17 target PII categories.

\subsubsection{LLMs using Basic, Math-Aware, and Segment-Aware Prompts}
    \label{sec:llm}

We evaluated the performance of six LLMs---Claude 4.5 Haiku, Claude 4.5 Opus, GPT-4 Turbo, GPT-5.2, Gemini 2.5 Flash, and Gemini 3 Pro---on the task of PII detection to examine the robustness of LLM-based methods. These models were selected because they are widely used and span lightweight, general-purpose, and high-capability model regimes. To ensure data security and compliance, all API requests were routed through Cornell University's LLM Gateway, LiteLLM\footnote{https://docs.litellm.ai/}. This centralized gateway provides a unified interface while enforcing strict privacy protocols; crucially, it operates under a zero-retention policy where payload data is neither logged by the gateway nor used for model training, aligning with institutional IRB requirements.

We first compared model performance under basic prompting against the baseline and selected the higher-performing model from each provider for subsequent evaluation with the math-aware and segment-aware prompting strategies, in order to assess whether these refinements further improve PII detection performance. The three prompt variants are described below.

\textbf{Basic Prompt}: A straightforward instruction acting as a PII specialist. It lists the 17 PII categories and requests the model to identify ``ALL PII'' in the message, returning a JSON list of text spans and types (exact text here\footnote{see 4 in https://github.com/ZhuqianZhou/MathEd-PII/blob/69012c0dbde3b3976f36533e2e44f5a5bfdfd718/ EDM2026\_paper\_183\_Appendix.pdf}).

\textbf{Math-Aware Prompt}: A refined version of the basic prompt that explicitly addresses the domain challenge. This prompt aims to reduce false positives stemming from numeric ambiguity by leveraging the model's general knowledge about math tutoring (exact text here\footnote{see 5 in https://github.com/ZhuqianZhou/MathEd-PII/blob/69012c0dbde3b3976f36533e2e44f5a5bfdfd718/ EDM2026\_paper\_183\_Appendix.pdf}).

\textbf{Segment-Aware Prompt}: A refined version of the basic prompt that explicitly points to the math segment label for each message. This prompt aims to reduce false positives stemming from numeric ambiguity through explicit contextual grounding (exact text in here\footnote{see 6 in https://github.com/ZhuqianZhou/MathEd-PII/blob/69012c0dbde3b3976f36533e2e44f5a5bfdfd718/ EDM2026\_paper\_183\_Appendix.pdf}).

\subsubsection{Evaluation Metrics and Statistical Analysis}
We evaluated PII detection performance using standard metrics: Precision, Recall, and the F1-score. Precision measures the proportion of detected text spans that correctly match the ground truth, Recall indicates the proportion of actual PII spans successfully identified, and the F1-score represents their harmonic mean. To account for the inherent variability and potential clustering of errors within different tutoring sessions, we calculated 95\% confidence intervals (CIs) using a bootstrap resampling method. We performed 1,000 bootstrap iterations by resampling tutoring transcripts (the primary unit of our corpus) with replacement. For each iteration, metrics were aggregated across the resampled sessions. The 95\% CIs reported in our results correspond to the 2.5th and 97.5th percentiles of the resulting bootstrap distribution.

\subsection{PII Detection Performance}
\subsubsection{Overall Performance}
Table~\ref{tab:results} summarizes the performance of Microsoft Presidio (baseline) and various LLMs using three types of prompts.

\begin{table}[h]
\centering
\caption{\textbf{PII Detection Performance Metrics (with 95\% Bootstrap Confidence Intervals)}}
\label{tab:results}
\resizebox{\columnwidth}{!}{%
\begin{tabular}{l c c c}
\toprule
\textbf{\large{Model}} & \textbf{\large{Precision}} & \textbf{\large{Recall}} & \textbf{\large{F1}} \\ 
\midrule
\textit{Baseline} & & & \\
\large{Presidio (Large)} & \begin{tabular} {c} \large{0.254} \\ \scriptsize{[0.225, 0.285]} \end{tabular} & \begin{tabular} {c} \large{0.747} \\ \scriptsize{[0.694, 0.794]} \end{tabular} & \begin{tabular} {c} \large{0.379} \\ \scriptsize{[0.344, 0.415]} \end{tabular} \\
\large{Presidio (Transformer)} & \begin{tabular} {c} \large{0.230} \\ \scriptsize{[0.208, 0.253]} \end{tabular} & \begin{tabular} {c} \large{\textbf{0.781}} \\ \scriptsize{[0.731, 0.826]} \end{tabular} & \begin{tabular} {c} \large{0.355} \\ \scriptsize{[0.328, 0.385]} \end{tabular} \\
\midrule \textit{Basic Prompting} & & & \\
\large{Claude 4.5 Haiku} & \begin{tabular} {c} \large{0.831} \\ \scriptsize{[0.802, 0.860]} \end{tabular} & \begin{tabular} {c} \large{0.620} \\ \scriptsize{[0.565, 0.675]} \end{tabular} & \begin{tabular} {c} \large{0.710} \\ \scriptsize{[0.670, 0.752]} \end{tabular} \\
\large{Claude 4.5 Opus} & \begin{tabular} {c} \large{0.911} \\ \scriptsize{[0.861, 0.946]} \end{tabular} & \begin{tabular} {c} \large{0.730} \\ \scriptsize{[0.677, 0.783]} \end{tabular} & \begin{tabular} {c} \large{0.810} \\ \scriptsize{[0.762, 0.853]} \end{tabular} \\
\large{Gemini 2.5 Flash} & \begin{tabular} {c} \large{0.848} \\ \scriptsize{[0.806, 0.887]} \end{tabular} & \begin{tabular} {c} \large{0.771} \\ \scriptsize{[0.723, 0.818]} \end{tabular} & \begin{tabular} {c} \large{0.807} \\ \scriptsize{[0.769, 0.844]} \end{tabular} \\
\large{Gemini 3 Pro} & \begin{tabular} {c} \large{0.887} \\ \scriptsize{[0.863, 0.910]} \end{tabular} & \begin{tabular} {c} \large{0.760} \\ \scriptsize{[0.706, 0.812]} \end{tabular} & \begin{tabular} {c} \large{0.818} \\ \scriptsize{[0.781, 0.851]} \end{tabular} \\
\large{GPT-4 Turbo} & \begin{tabular} {c} \large{0.732} \\ \scriptsize{[0.695, 0.769]} \end{tabular} & \begin{tabular} {c} \large{0.578} \\ \scriptsize{[0.530, 0.630]} \end{tabular} & \begin{tabular} {c} \large{0.646} \\ \scriptsize{[0.609, 0.685]} \end{tabular} \\
\large{GPT-5.2} & \begin{tabular} {c} \large{0.759} \\ \scriptsize{[0.723, 0.795]} \end{tabular} & \begin{tabular} {c} \large{0.707} \\ \scriptsize{[0.656, 0.759]} \end{tabular} & \begin{tabular} {c} \large{0.732} \\ \scriptsize{[0.695, 0.767]} \end{tabular} \\
\midrule \textit{Math-Aware Prompting} & & & \\
\large{Claude 4.5 Opus} & \begin{tabular} {c} \large{0.905} \\ \scriptsize{[0.860, 0.938]} \end{tabular} & \begin{tabular} {c} \large{0.720} \\ \scriptsize{[0.661, 0.769]} \end{tabular} & \begin{tabular} {c} \large{0.802} \\ \scriptsize{[0.756, 0.842]} \end{tabular} \\
\large{Gemini 3 Pro} & \begin{tabular} {c} \large{0.869} \\ \scriptsize{[0.826, 0.906]} \end{tabular} & \begin{tabular} {c} \large{0.735} \\ \scriptsize{[0.685, 0.787]} \end{tabular} & \begin{tabular} {c} \large{0.796} \\ \scriptsize{[0.756, 0.835]} \end{tabular} \\
\large{GPT-5.2} & \begin{tabular} {c} \large{0.774} \\ \scriptsize{[0.740, 0.809]} \end{tabular} & \begin{tabular} {c} \large{0.721} \\ \scriptsize{[0.670, 0.764]} \end{tabular} & \begin{tabular} {c} \large{0.747} \\ \scriptsize{[0.711, 0.780]} \end{tabular} \\
\midrule \textit{Segment-Aware Prompting} & & & \\
\large{Claude 4.5 Opus} & \begin{tabular} {c} \large{\textbf{0.934}} \\ \scriptsize{[0.913, 0.951]} \end{tabular} & \begin{tabular} {c} \large{0.730} \\ \scriptsize{[0.682, 0.784]} \end{tabular} & \begin{tabular} {c} \large{0.820} \\ \scriptsize{[0.785, 0.855]} \end{tabular} \\
\large{Gemini 3 Pro} & \begin{tabular} {c} \large{0.888} \\ \scriptsize{[0.862, 0.913]} \end{tabular} & \begin{tabular} {c} \large{0.764} \\ \scriptsize{[0.713, 0.811]} \end{tabular} & \begin{tabular} {c} \large{\textbf{0.821}} \\ \scriptsize{[0.788, 0.851]} \end{tabular} \\
\large{GPT-5.2} & \begin{tabular} {c} \large{0.800} \\ \scriptsize{[0.760, 0.838]} \end{tabular} & \begin{tabular} {c} \large{0.720} \\ \scriptsize{[0.670, 0.766]} \end{tabular} & \begin{tabular} {c} \large{0.758} \\ \scriptsize{[0.720, 0.793]} \end{tabular} \\
\bottomrule
\end{tabular}%
}
\end{table}

The baseline Presidio models achieved high recall (with the Transformer version reaching a point estimate of 78.1\%), but suffered from exceptionally poor precision (spanning 0.23--0.25). This confirms that generic PII detection tools tend to over-redact math-rich content by failing to distinguish mathematical notation from sensitive identifiers. While the point estimates for Presidio's recall appear high, several top-performing LLMs achieved recall levels that are statistically comparable (with overlapping 95\% bootstrap confidence intervals) while providing vastly superior precision and F1 scores. Based on these initial results, Claude 4.5 Opus (Anthropic), Gemini 3 Pro (Google), and GPT-5.2 (OpenAI) were selected as the primary focus for evaluating domain-specific prompt refinements.

Math-aware prompting generally improved the precision of the selected models compared to the basic prompt across most configurations. Segment-aware prompting, which explicitly grounds the detection process in the mathematical context of the message, yielded the highest overall performance for all three models. Notably, Claude 4.5 Opus achieved the highest precision (0.934, 95\% CI [0.913, 0.951]) under this condition, while Gemini 3 Pro achieved the highest overall F1 score (0.821, 95\% CI [0.788, 0.851]). Crucially, the segment-aware versions of these models maintain recall levels that are on par with the baseline, while reducing false positives by as much as 90\%. These results demonstrate that incorporating mathematical discourse context is essential for minimizing over-redaction.

\subsubsection{Performance by PII Categories}
Given that both the baseline and the top-performing LLMs achieved comparable recall levels (as discussed in the previous section), we zoom in on category-level performance to investigate how domain-aware prompting reduces false positives to achieve superior precision. Table~\ref{tab:per_type} provides a breakdown of the specific sources of over-redaction, focusing on False Positives (FP) and Precision across the target PII categories under two model configurations: the highest-performing baseline, Presidio LG (using the \texttt{en\_core\_web\_lg} SpaCy NER model), and the highest-performing LLM configuration, Gemini 3 Pro with segment-aware prompting.

\begin{table}[h]
\centering
\caption{\textbf{PII Detection False Positives and Precision by Category (ordered by Precision from the highest to the lowest from Gemini 3 Pro with Segment Prompting)}}
\label{tab:per_type}
\renewcommand{\arraystretch}{1.5}
\resizebox{\columnwidth}{!}{%
\begin{tabular}{l c | c c | c c}
\toprule 
\textbf{\large{PII Type}} & \textbf{\large{Total}} & \multicolumn{2}{c|}{\textbf{\large{Presidio LG}}} & \multicolumn{2}{c}{\textbf{\large{Gemini 3 Pro Seg.}}} \\
 & & \textbf{\large{FP}} & \textbf{\large{Prec}} & \textbf{\large{FP}} & \textbf{\large{Prec}} \\ \midrule
 \large{US\_DRIVER\_LICENSE} & \large{2} & \large{344} & \large{0.000} & \large{0} & \large{1.000} \\
\large{PERSON} & \large{1422} & \large{330} & \large{0.775} & \large{63} & \large{0.949} \\
\large{LOCATION} & \large{121} & \large{153} & \large{0.349} & \large{5} & \large{0.896} \\
\large{NRP} & \large{25} & \large{45} & \large{0.286} & \large{2} & \large{0.857} \\
\large{URL} & \large{187} & \large{21} & \large{0.859} & \large{21} & \large{0.852} \\
\large{SCHOOL} & \large{73} & \large{17} & \large{0.638} & \large{10} & \large{0.831} \\
\large{GRADE\_LEVEL} & \large{103} & \large{2} & \large{0.973} & \large{22} & \large{0.750} \\
\large{IP\_ADDRESS} & \large{2} & \large{0} & \large{1.000} & \large{1} & \large{0.667} \\
\large{PHONE\_NUMBER} & \large{2} & \large{6} & \large{0.250} & \large{1} & \large{0.667} \\
\large{AGE} & \large{8} & \large{3} & \large{0.000} & \large{6} & \large{0.538} \\
\large{COURSE\_NUMBER} & \large{40} & \large{46} & \large{0.193} & \large{43} & \large{0.411} \\
\large{DATE} & \large{3} & \large{3386} & \large{0.001} & \large{16} & \large{0.059} \\
\large{US\_BANK\_NUMBER} & \large{0} & \large{5} & \large{0.000} & \large{-} & \large{-} \\
\bottomrule
\end{tabular}%
}
\end{table}

The most striking improvement is observed in numeric categories where the baseline is particularly prone to over-redact mathematical entities. For instance, Presidio generated 3,386 false positives for \texttt{DATE} and 344 for \texttt{US\_DRIVER\_LICENSE}, confirming that rule-based systems frequently misinterpret math spans as structured identifiers. In contrast, Gemini 3 Pro with segment-aware prompting reduced \texttt{DATE} false positives by over 99.5\% (down to 16) and eliminated false positives for \texttt{US\_DRIVER\_LICENSE} entirely. Similarly, for context-heavy categories like \texttt{PERSON} and \texttt{LOCATION}, the LLM significantly improved precision by reducing the number of false detections. These results underscore that the superior performance of domain-aware LLMs is primarily driven by their ability to selectively filter mathematical noise while maintaining a capture rate on par with standard tools.

\subsubsection{Performance by Segment Types}
To further investigate how well the LLM addressed numeric ambiguity, we evaluated model performance separately for messages labeled as NON-MATH versus MATH conversational segments. Table~\ref{tab:segment_results} reveals a consistent performance gap across the models, with precision being systematically lower in the MATH segment compared to the NON-MATH segment.

\begin{table}[h]
\centering
\caption{\textbf{PII Detection Performance by Segment Type (NON-MATH vs. MATH)}}
\label{tab:segment_results}
\renewcommand{\arraystretch}{1.5}
\resizebox{\columnwidth}{!}{%
\begin{tabular}{l c | c c c | c c c}
\toprule
\textbf{\large{Model}} & \textbf{\large{Prompt}} & \multicolumn{3}{c|}{\textbf{\large{NON-MATH Segments}}} & \multicolumn{3}{c}{\textbf{\large{MATH Segments}}} \\
 & & \textbf{\large{Prec}} & \textbf{\large{Rec}} & \textbf{\large{F1}} & \textbf{\large{Prec}} & \textbf{\large{Rec}} & \textbf{\large{F1}} \\ \midrule
\large{Presidio LG} & \large{-} & \large{0.346} & \large{0.756} & \large{0.475} & \large{0.142} & \large{0.720} & \large{0.237} \\
\large{Gemini 3 Pro} & \large{Basic} & \large{0.913} & \large{0.819} & \large{0.864} & \large{0.800} & \large{0.591} & \large{0.680} \\
\large{Gemini 3 Pro} & \large{Segment} & \large{0.909} & \large{0.815} & \large{0.860} & \large{0.819} & \large{0.618} & \large{0.705} \\
\bottomrule
\end{tabular}%
}
\end{table}

The stratified analysis in Table~\ref{tab:segment_results} provides strong empirical evidence for the numeric ambiguity hypothesis and demonstrates the effectiveness of segment-aware prompting in mitigating it. The baseline Presidio LG model exhibits a big drop in precision when moving from NON-MATH (0.346) to MATH segments (0.142), confirming that generic rule-based systems frequently misinterpret mathematical notation as sensitive identifiers. This 59\% decline in precision within math-dense regions highlights the risk of over-redaction in tutoring transcripts when domain context is ignored.

In contrast, the Gemini 3 Pro model maintains substantial precision in both contexts, though a performance gap between segments remains. Crucially, the introduction of segment-aware prompting specifically improves the model's ability to navigate numeric ambiguity within mathematical discourse. In MATH segments, segment-aware prompting increases both precision (0.800 to 0.819) and recall (0.591 to 0.618) compared to basic prompting. This simultaneous gain in both metrics is particularly noteworthy; it suggests that providing the model with an explicit context label allows it to adopt a more nuanced decision boundary—correctly identifying PII that it might otherwise miss (higher recall) while simultaneously avoiding false detections triggered by mathematical symbols (higher precision). The negligible difference in performance within NON-MATH segments further indicates that the benefits of this strategy are structurally concentrated in the regions where numeric ambiguity is most severe.

\section{Discussion and Conclusion}

This study investigated the challenge of utility-preserving de-identification in the context of math tutoring transcripts, focusing on the phenomenon of numeric ambiguity. By introducing MathEd-PII, the first benchmark dataset for this domain, we provided a rigorous foundation for evaluating PII detection methods that balance privacy protection with instructional fidelity.

Our findings provide strong empirical evidence that numeric ambiguity is a structural failure mode in standard PII detection systems. Generic tools like Microsoft Presidio often misinterpret mathematical expressions as sensitive identifiers, such as dates or driver's license numbers. Our density-based segmentation analysis revealed that over half of all false positives (55.5\%) were concentrated in math-dense regions, which comprise only about a third of the corpus tokens. In these regions, the ratio of false positives to true PII reached nearly 5:1, highlighting a significant risk of removing core instructional content that is essential for educational research.

We demonstrated that LLMs, particularly when guided by domain-aware prompting strategies, significantly outperform traditional baselines. While Presidio achieved a respectable recall, its low precision (F1: 0.380) renders it problematic for creating high-utility datasets. In contrast, Gemini 3 Pro with segment-aware prompting achieved an F1 of 0.821, maintaining high recall while reducing false positives by over 90\% in critical numeric categories. The success of segment-aware prompting suggests that providing models with explicit contextual grounding---identifying whether a message is likely to be mathematical---allows them to adopt more nuanced decision boundaries that better distinguish task-content from personal disclosure.

The creation of MathEd-PII through a human-in-the-loop LLM surrogation workflow also offers a methodological contribution. It demonstrates how high-quality benchmark data-sets can be generated even when original, unredacted data cannot be directly shared, by leveraging the generative capabilities of LLMs to create privacy-preserving surrogates that maintain discourse coherence.

However, several limitations and avenues for future work remain. First, while MathEd-PII is the first math-focused PII surrogate benchmark, it is based on 1,000 sessions from a single U.S.-based tutoring platform. As a result, the generalization behavior of the proposed methodology across broader educational domains remains an open question. Expanding this benchmark to include more diverse curricula, grade levels, instructional settings, and international contexts would improve the generalizability of our findings and enable more systematic evaluation across heterogeneous datasets. More broadly, our approach to ensuring applicability across different contexts is to emphasize domain-aware prompting principles that are adaptable to new educational settings, while future work will focus on validating these strategies on additional datasets and subject areas beyond mathematics tutoring.

Second, while this work focused primarily on addressing the precision issues caused by numeric ambiguity, further optimizing recall remains an important avenue for future exploration. Although our domain-aware prompting strategies yielded substantial improvements in F1 and maintained recall levels comparable to standard baselines, a portion of PII still remains undetected. While a detailed investigation into maximizing recall was beyond the scope of this paper, there are several promising directions to bridge this gap, such as: (1) developing multi-stage ``cascade'' architectures that combine high-recall rule-based systems with high-precision LLM verification; (2) exploring advanced prompting techniques like multi-agent debate to enhance sensitivity to subtle PII mentions; and (3) leveraging the MathEd-PII dataset to fine-tune specialized models that capture domain-specific patterns more effectively and robustly than prompt-only approaches. In addition, the current methodology relies entirely on prompting large language models, which may be brittle across model versions, prompting styles, or unseen distributions. Fine-tuning specialized domain-adapted models therefore represents an important future direction for improving robustness, consistency, and transferability across datasets and deployment settings. Finally, future work could also investigate making these solutions more efficient for low-cost, on-device deployment at educational institutions.

In conclusion, as dialogue-based interaction data becomes increasingly available to educational data mining community, the need for domain-aware de-identification becomes paramount. This work shows that by incorporating mathematical context into the de-identification pipeline, we can move closer to the goal of ``utility-preserving privacy''---protec-ting student and tutor identities without sacrificing the very data that makes educational research possible.

\section{Acknowledgments}
This material is based upon initial work completed under National Science Foundation Grant No. 2321499, and support from the Gates Foundation and the Chan Zuckerberg Initiative. Any opinions expressed in this material are those of the authors and do not necessarily reflect funders' views.

\section{Data Availability}
The MathEd-PII dataset is publicly available at  https://huggingface.co/datasets/NationalTutoringObservatory\newline /MathEd-PII.

\bibliographystyle{abbrv}
\bibliography{manuscript}

\end{document}